\documentclass[final,3p,times,twocolumn]{elsarticle}

\usepackage{amssymb}
\usepackage[colorlinks=true]{hyperref}
\usepackage{graphicx}
\usepackage{subcaption}
\usepackage{lscape}
\usepackage{amsmath}
\usepackage{multirow}
\usepackage{xcolor} 
\usepackage{caption}
\usepackage{adjustbox}
\usepackage[figuresright]{rotating}
\usepackage{algorithm} 
\usepackage{algpseudocode} 
\usepackage{comment}
\usepackage{hyperref}
\usepackage{booktabs} 
\usepackage{rotating} 

\usepackage{natbib}
\setcitestyle{authoryear,open={(},close={)},citesep={;}} 

\usepackage{calc}
\makeatletter
\def\ps@pprintTitle{%
  \let\@oddhead\@empty
  \let\@evenhead\@empty
  \def\@oddfoot{\reset@font\hfil\thepage\hfil}
  \let\@evenfoot\@oddfoot
}
\makeatother

\begin{document}

\begin{frontmatter}

\title{A Comparative Benchmark of Real-time Detectors for Blueberry Detection towards Precision Orchard Management}

\author[label1]{Xinyang Mu}
\author[label1]{Yuzhen Lu}
\author[label1]{Boyang Deng}

\address{Yuzhen Lu (luyuzhen@msu.edu) is the corresponding author}
\address[label1]{Department of Biosystems and Agricultural Engineering, Michigan State University, East Lansing, MI 48824, USA}

\begin{abstract}
Computer vision powered by artificial intelligence offers a promising tool for blueberry growers to automate orchard tasks such as harvest maturity assessment and yield estimation. However, blueberry detection in natural environments remains challenging due to variable natural lighting, occlusions by leaves and branches, and motion blur due to environmental factors and imaging devices. Deep learning-based object detectors promise to address these challenges, but they are data-driven, demanding a large-scale, diverse dataset that captures the real-world complexities. Moreover, deploying these models in practical scenarios with limited computing resources requires the right accuracy/speed/memory trade-off in model selection. This study presents a novel comparative benchmark analysis of advanced real-time object detectors, including YOLO (You Only Look Once) (v8-v12) and RT-DETR (Real-Time Detection Transformers) (v1-v2) families, consisting of 36 model variants, evaluated on a newly curated dataset for blueberry detection. This dataset comprises 661 canopy images collected with smartphones during the 2022-2023 seasons, consisting of 85,879 manually annotated instances (including 36,256 ripe and 49,623 unripe blueberries) across a wide range of lighting conditions, occlusions, and fruit maturity stages. Among the YOLO models, YOLOv12m achieved the best accuracy with a mAP@50 of 93.3\%, while RT-DETRv2-X obtained a mAP@50 of 93.6\%, the highest among all the RT-DETR variants. The inference time varied with the model scale and complexity, and the mid-sized models appeared to offer a good balance between accuracy and speed. To further enhance detection performance, all the models were fine-tuned using Unbiased Mean Teacher-based semi-supervised learning (SSL) on a separate set of 1,035 unlabeled images acquired by a ground-based machine vision platform in 2024. This resulted in accuracy gains ranging from -1.4\% to 2.9\%, with RT-DETR-v2-X achieving the best mAP@50 of 94.8\%. More in-depth research into SSL is needed to better leverage cross-domain unlabeled data. Both the dataset and software programs developed in this study are made publicly available ({\url{https://github.com/rogermu789/BlueberryBenchmark}}) to support further research.
\end{abstract}


\end{frontmatter}

\section{Introduction}
\label{sec:intro}

Blueberries have become an increasingly valuable crop, marked by their rich antioxidant content and diverse nutritional benefits, including the potential to lower the risks of cardiovascular disease, cancer, and other aging-related ailments \citep{seeram2008berry,johnson2015daily,basu2010blueberries,krikorian2022blueberry}. Growing consumer awareness of these health benefits has driven a surge in global demand and cultivation. In the United States, blueberries were grown on over 103,000 acres in 2023, generating a farmgate value exceeding \$1 billion \citep{usda2024noncitrus} and reinforcing the nation’s status as a leading producer and consumer \citep{usda2019blueberries}. The U.S. industry primarily cultivates three blueberry varieties: highbush, lowbush, and rabbiteye, with highbush blueberries dominating commercial production \citep{retamales2018blueberries}. This rapid expansion of the blueberry market has spurred interest in leveraging modern technologies to optimize orchard practices.

As blueberry production continues to grow, there is an increasing need for computer vision techniques, particularly in the areas of harvest maturity assessment and yield estimation, to support effective crop management. Manual assessment of blueberry maturity based on the fruit skin color is the current practice for harvest decision-making. In the early stage, the blueberry fruit color is green, gradually transitions to pink, red, and eventually turns deep blue or black as the fruit reaches full maturity. The first-hand-picking event commonly takes place when about 20\% of the fruit on a bush is blue, and repeated picking will happen as the fruit maturity advances, while harvesting can be considered when 50–60\% of the crop is blue \citep{msu2025growth,devetter2022harvesting}. Manual assessment of the fruit harvest maturity, which requires counting fruit in different colors (e.g., green, blue) is an extremely time-consuming, labor-intensive process that is only feasible for a small number of selected branches. The manual approach often results in inconsistent and inaccurate assessments that can compromise harvest timing and the quality of the harvested fruit. Currently, growers often guess the crop yield based on the yield data from previous seasons, as there are no appropriate in-season yield estimation methods or tools available for blueberry growers. Computer vision technology promises to provide automated or high-throughput methods for blueberry detection and maturity assessment, thereby facilitating harvest decision-making and yield estimation.

In real orchards, the detection of blueberries presents a range of challenges that could impact the performance of computer vision systems. This is largely because of the fact that blueberries have small fruit sizes, and are often densely packed and clustered or occluded by the leaves and twigs \citep{deng2024detection}, complicating accurate detection of individual fruit instances. Other confounding factors include the variation in natural lighting conditions and an unstructured background. Images captured in orchards are subject to fluctuating sunlight, intermittent shadows, and dynamic cloud cover, resulting in inconsistent illumination and color distortions that complicate the detection of subtle indicators of fruit ripeness \citep{bargoti2017deep}. Environmental influences such as wind-induced motion and camera shaking further contribute to image artifacts such as motion blur, thereby degrading image clarity and reducing precision \citep{fergus2007weakly,whyte2012non}. Seasonal variations and different growth stages add yet another layer of complexity, as the visual color of blueberries can vary significantly over time. These multifaceted challenges remain to be addressed before computer vision technology can be deployed successfully in orchard conditions for blueberry detection.  

Recent advancements in deep learning have inspired efforts in computer vision for blueberry detection using canopy images acquired in real orchard environments. The YOLO (You Only Look Once) family, among real-time convolutional neural network (CNNs), has been particularly influential in agriculture \citep{sapkota2025rf,badgujar2024agricultural}. Several studies have been carried out using YOLO detectors for the detection of blueberries of varied maturity. \cite{schumann2019detection} applied four versions of YOLOv3 for detecting wild blueberries of three levels of maturity, achieving the mAP@50 (mean average precision at an intersection-over-union threshold of 50\%) of 85.3 \% with inference times of 28 ms. \cite{maceachern2023detection} reported on using YOLOv4 with mAP@50 of 79.8 \% and 88.1 \% in detecting 2-class and 3-class blueberries, respectively, as well as the mean absolute error of 24.1 \% in fruit detection-based yield estimation. \cite{liu2023blueberry} proposed BlueberryYOLO based on YOLOv5 for enhanced fruit detection, achieving a mAP@50 of 78.3\% in detecting blueberries of three levels of ripeness at 47 FPS (frames per second) on an edge computing unit. \cite{li2023blueberry} applied YOLOv8 for detecting blueberries from multi-view images (i.e., top, left, and right views) of canopies, obtaining a mAP@50 of 77.3 \% based on YOLOv8x. Most of these efforts, however, experimented with a small number of images captured for localized blueberry bushes without replicated testing, which may undermine the credibility of model performance, and additionally, the datasets reported in all these studies were not made publicly available. In addition to CNN-based detectors, vision transformers, especially real-time detection transformers (RT-DETRs), have emerged as competitive alternatives for computer vision tasks in precision agriculture \citep{mehdipour2025vision}. Recent research has shown the remarkable performance of RT-DETRs comparable to or surpassing YOLOs in scenarios with complex backgrounds and varying illumination \citep{aguilera2023comprehensive,chen2024rt,sapkota2025rf}, but RT-DETRs remain to be fully assessed for blueberry detection, maturity assessment, and yield estimation.

Despite the progress in blueberry detection using deep learning, the available datasets remain limited in size and diversity, which limits the development of robust, practically applicable models. \cite{deng2024detection} presented the first publicly available dataset acquired using smartphones for blueberry detection. The data was released in the Zenodo repository \citep{lu2024blueberrydcm}, consists of 140 canopy images with 17,955 instances of two classes (ripe and unripe fruit) acquired using handheld cameras in both research farms and commerical orchards from different locations. Building on the dataset, an iOS-based mobile app (BlueberryCounter) with YOLOv8-based fruit detectors deployed was developed as a handy tool for blueberry growers \citep{deng2025development}. Although this dataset offers an important testbed for model prototyping and evaluation, it is still short of capturing the full range of variability found in the natural environment, such as diverse lighting conditions, occlusions, and multiple stages of fruit maturity. Recently, \cite{li2025field} reported a blueberry fruit detection dataset alongside datasets for other visual tasks. The detection dataset consisted of labelled 405 images with over 100,000 berries captured from blueberry plants (mostly young lowbush) in varied heights by platform-based and handheld imaging devices in the university-managed research field for blueberry breeding \citep{li2023blueberry}. The authors shared both their software programs and four datasets for different tasks in Kaggle \citep{li2025field}. There remains a pressing need for creating larger, more diverse datasets that better reflect the complexities of diverse real-world orchard conditions, ultimately boosting the performance and reliability of deep learning models for practical application.

Compared to imaging processes, the manual annotation of acquired imagery for fruit detection is labor- and resource-intensive and can be a real bottleneck in dataset creation. Annotating blueberry canopy images is particularly challenging due to the high density of fruit, their small size, and frequent occlusions, all of which demand detailed, instance-level labeling by trained personnel. Semi-supervised learning (SSL) offers a potential solution to this challenge by incorporating large volumes of unlabeled data into the training process \citep{yang2022survey}, thereby reducing the need for manual annotations. For instance, \cite{ciarfuglia2023weakly} applied weakly and semi-supervised techniques to the detection, segmentation, and tracking of table grapes using limited and noisy data, showing that robust performance could be achieved even when the amount of labeled data was minimal. \cite{johanson2024s3ad} introduced S³AD, a semi-supervised system for small apple detection in orchard environments that utilizes both labeled and unlabeled images to improve detection performance compared to fully supervised baselines while significantly cutting down on annotation efforts. Semi-supervised techniques in a teacher-student framework have proven effective in leveraging unlabeled data for object detection tasks \citep{wang2023dual}, highlighting their potential to address the shortage of labeled data in agricultural applications. Given the successes in related fruit detection tasks, semi-supervised learning is worthy of investigation for enhancing blueberry detection in complex orchard environments.

This study contributed to three specific objectives. It was aimed to: 1) present the largest publicly available dataset that captures the variability and complexity of real orchard environments for blueberry detection and maturity assessment, which comprise 661 images with 85,879 annotated instances of ripe and unripe fruit, alongside an unlabeled set of 1,035 images acquired by a machine vision platform, 2) conduct a comparative evaluation of a large suite of state-of-the-art real-time deep learning detectors, including five latest YOLO versions (v8-v12) and two versions of RT-DETRs (v1-v2) at varied scales, for blueberry detection, and 3) evaluate the efficacy of SSL techniques for reducing the dependence on labelled data process and improving the accuracy of supervised detectors. These contributions are expected to enhance blueberry detection performance, facilitate more precise maturity assessment and yield estimation, and provide a valuable resource for future research in optimizing computer vision-enabled blueberry orchard management.

\section{Materials and Methods}
\label{sec:data_collection}
\subsection{Blueberry Dataset}
\label{sec:2.1}
\begin{figure*}[!ht]
 \centering
 \includegraphics[width=1.0\textwidth]{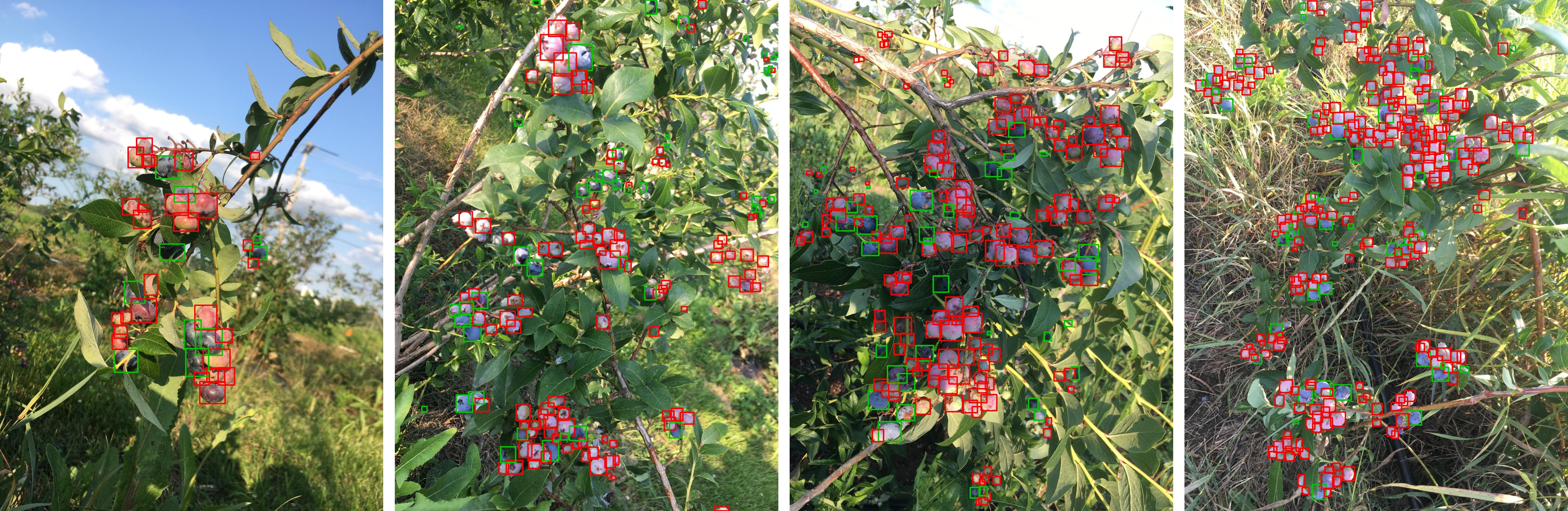}
 \caption{Examples of labelled blueberry images across diverse field conditions, including variations in fruit size, camera angle, and illumination. The red and green bounding boxes indicate unripe and ripe berries, respectively. }
 \label{fig:fig1}
\end{figure*}

The dataset in this study contains two sets of labelled and unlabeled images of blueberry canopies. The labelled dataset, which was used for blueberry detection based on supervised learning, consists of 661 images captured with different smartphones (i.e., iPhone SE, and iPhone v11, v12 and v13); among the labelled data, a set of 140 images was captured in 2022 and detailed in prior work \citep{deng2024detection}, while the remaining 521 images were taken in 2023, with 161 images captured in a commercial blueberry farm (Rockford, MI) and the rest on a research farm (Holt, MI) of Michigan State University (MSU). These images were collected at different spatial scales, ranging from individual branches to entire bush canopies, and under diverse natural lighting conditions, making the dataset suitable for developing robust blueberry detection models.

The dataset for supervised learning was manually labelled by trained personnel following a standardized annotation protocol. The annotation was performed using the VGG Image Annotator (version 2.0.12) . During the annotation process, each visible blueberry fruit was classified as either “Blue” or “Unblue” using a checkbox, corresponding to ripe and unripe fruit based on its skin color, varying from green to dark blue. Given the small size of the blueberries, annotators zoomed in on each image (300\% or more) during the process. Additionally, every annotated image underwent a quality review by independent personnel to ensure accuracy before being added to the final dataset. The exported annotation files were converted to YOLO-format text files and COCO-format JSON files  compatible with the requirements of YOLO and RT-DETR models, respectively, for blueberry detection. Table~\ref{tab:table1} summarizes the statistics of the labelled dataset. Examples of the images with labelled blueberry instances are shown in Figure~\ref{fig:fig1}.

\begin{table}[ht]
\caption{Statistics of the labelled blueberry dataset: the number (\#) of images, annotated bounding boxes, and counts of ripe and unripe fruits collected during 2022 and 2023.}
\label{tab:table1}
    \centering
    \resizebox{0.5\textwidth}{!}{%
        \begin{tabular}{lcccc}
            \toprule
            Year & \# of Images & \# of Bounding Boxes & \# of Ripe Fruits & \# of Unripe Fruits \\
            \midrule
            2022     & 140                    &   17,854                     & 6,967  &10,887\\
            2023       & 521                   & 68,025                       & 29,289  &38,736\\
            Total     & 661                    & 85,879                       & 36,256  &49,623\\
            \bottomrule
        \end{tabular}
    }
\end{table}

\begin{figure*}[!ht]
 \centering
 \includegraphics[width=1.0\textwidth]{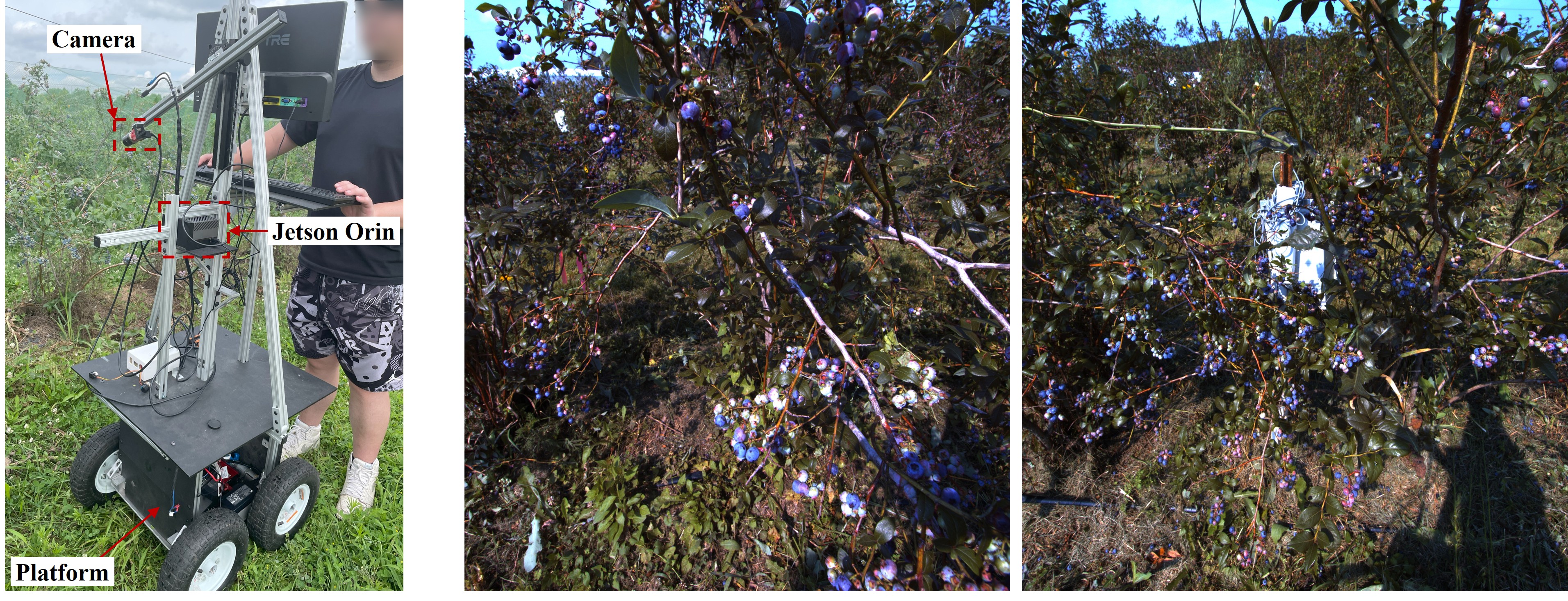}
 \caption{A ground-based mobile platform for scanning blueberry bushes (left), and example unlabeled images of highbush blueberries captured in the season of 2024.}
 \label{fig:fig2}
\end{figure*}

In addition to the labelled data above, a new set of data was acquired in the season of 2024 using a machine vision camera (Alvium 1800 U-811C, Allied Vision, Stadtroda, Germany) attached with a 5-mm focusing lens (Kowa, Nagoya, Japan), which was mounted on a lightweight ground-based mobile platform, as shown in Figure~\ref{fig:fig2}. The platform, equipped with a Jetson Orin computer (NVIDIA, Santa Clara, CA), scanned blueberry bushes at a traveling speed of approximately 0.2 m/s from a side view on the MSU blueberry farm (Hort, MI). A custom-written software program was implemented for automatically capturing an image every 3 seconds at a resolution of 2848×2848 pixels. A total of 1035 images acquired by the platform were used in this study to exploit SSL-based blueberry detection. It is noted that the platform dataset had been annotated at the time of writing, consisting of 65,967 blueberry instances, but it was treated as unlabeled by ignoring the annotations for SSL-based blueberry detection in this study. 

Figure~\ref{fig:fig2} shows examples of the images acquired by the platform.  Compared to the labeled images by smartphones, the platform images give a wider view of blueberry bushes, including multiple clusters in a single image, significantly increasing the quantity and diversity of fruit instances. Although individual berries appear smaller and are often subject to occlusion, overlap, and glare from reflective foliage, the higher fruit density within each image provides richer contextual information for the learning process. The abundance and variability of the dataset can also be beneficial for robust blueberry detection in practical orchard conditions and enhanced model generalization. However, the platform dataset is visually more challenging due to dynamic scenes, reduced pixel resolution, and complex canopy backgrounds, which may pose challenges in the SSL process to generating high-quality pseudo labels by models trained the labeled dataset.

\subsection{Real-Time Detectors}
Real-time fruit detection is important to precision orchard management tasks that demand efficient, on-site decision-making and operations. Over the past decade, CCN-based single-stage detectors that perform bounding box prediction and classification simultaneously by a single network, requiring no separate step for generating regional proposals, have been widely used in various real-world applications. Among these deep networks, the YOLO series, a family of object detectors, is the most remarkable due to its good trade-off between speed and accuracy, and they are evolving readily, for which each mainstream variant addresses limitations for better performance. Alternatively, detection transformers (DETR) have gained increasing attention, and the RT-DETRs have emerged as a competitive option for real-time applications compared to YOLOs. Hence, the state-of-the-art YOLOs and RT-DETRs were selected for blueberry detection, given their potential for orchard applications.

\subsubsection{YOLO Object Detectors}
Recent developments in the YOLO series since 2023 have led to the evolution from YOLOv8 \citep{range2023brief} to YOLOv13 \citep{lei2025yolov13}. Each variant introduces distinct architectural designs and training strategies while addressing detection challenges such as small objects and occlusions, which can be implemented at different network scales for handling tasks of varied complexity. YOLOv8 represents an important milestone in the YOLO series due to a myriad of technological innovations. Featuring a redesigned backbone aimed at extracting fine-grained features, YOLOv8 is promising for detecting small objects such as blueberries \citep{deng2024detection}. The detector employs cross-stage partial connections to enhance gradient flow and reduce computational redundancy, leading to improved training stability. The network’s neck integrates a refined feature pyramid structure that effectively aggregates multi-scale information, ensuring that both local details and broader contextual cues are captured. Furthermore, the detection head in YOLOv8 introduces an anchor-free approach, simplifying model training and improving detection efficiency.

YOLOv9 \citep{wang2024yolov9} implements an advanced attention mechanism within its detection head to focus on critical features amidst cluttered backgrounds. Specifically, spatial and channel attention modules are integrated to recalibrate feature maps dynamically, which can help improve the discrimination of blueberry clusters from surrounding foliage. YOLOv9 also introduces a dynamic anchor box adjustment mechanism during training, allowing the model to fine-tune anchor parameters in response to the actual distribution of object sizes. This adaptation can enhance localization precision and overall detection performance, particularly in scenarios where targets such as blueberries are partially occluded or densely packed. YOLOv10 \citep{wang2024yolov10} represents another evolution in the YOLO series by integrating CNNs with transformer-based modules, creating a hybrid architecture that captures both local and long-range dependencies. This design enables richer semantic representation, which can be beneficial for resolving closely packed or overlapping blueberry instances. The model features an enhanced multi-scale feature fusion module that effectively combines detailed spatial information with global context, and it employs a refined non-maximum suppression strategy to improve the resolution of overlapping detections. Additionally, YOLOv10 benefits from state-of-the-art training techniques, including sophisticated data augmentation and optimized hyperparameters, thereby enabling potentially higher accuracy and real-time performance under challenging orchard conditions.

YOLOv11 \citep{jocher2025yolo11edge} builds upon the hybrid architecture of YOLOv10 by introducing a more efficient backbone designed through neural architecture search, which reduces computational overhead while maintaining high accuracy. This efficiency makes YOLOv11 well-suited for real-time applications in orchard environments, where hardware resources may be limited. The model also integrates refined knowledge distillation strategies, enabling compact versions of YOLOv11 to learn from larger teacher networks, thereby preserving detection accuracy while supporting faster inference. These advancements can enhance the ability to detect small and partially occluded blueberries with improved precision and robustness. YOLOv12 \citep{tian2025yolov12} represents the next leap in detection performance by incorporating adaptive transformer-based modules that dynamically allocate attention according to scene complexity. This allows the network to better distinguish overlapping blueberry instances and suppress background noise from leaves and branches. Furthermore, YOLOv12 introduces an improved multi-scale feature aggregation framework, enabling richer representation across varying object sizes. Combined with advanced data augmentation and optimized training schedules, YOLOv12 promises to deliver state-of-the-art detection performance in challenging orchard conditions, achieving higher reliability for dense and cluttered blueberry clusters. 

\subsubsection{Real-Time Detection Transformers (RT-DETR)} 
While the YOLO series has demonstrated notable efficiency and accuracy in real-time object detection, end-to-end detection transformers (DETRs), which employ self-attention mechanisms to capture long-range dependencies and global context, address some of the limitations of convolutional architectures and offer new avenues for improved handling of overlapping objects and complex backgrounds, which are critical challenges in applications like blueberry detection. RT-DETR implements innovations in network architectures and training strategies, overcoming the computational costs of previous DETR and extending it to real-time detection scenarios. Presented below is a brief overview of the major innovations of two advanced, real-time versions of DETR, i.e., RT-DETR-v1 \citep{zhao2024detrs} and RT-DETR-v2 \citep{lv2024rt}.

RT-DET-v1 is the first transformer-based framework adapted for real-time object detection. This model leverages an encoder-decoder architecture where the encoder captures global contextual relationships using multi-head self-attention, and the decoder refines object queries to produce final detections. To reduce computational costs, RT-DETR-v1 incorporates a new efficient hybrid encoder design that decouples inter-scale interaction and cross-scale fusion, expediting the processing of multi-scale features while maintaining competitive accuracy. The network also implements an uncertainty-minimal query selection scheme that provides high-quality encoder features, improving the accuracy of the detector. Additionally, the RT-DETR also supports flexible model scaling and speed adjustments to accommodate different scenarios of practical application. These features encourage the use of the detector as a competitive alternative to YOLOs for blueberry detection.

Building upon the framework of RT-DETR-v1, RT-DETR-v2 introduces several enhancements aimed at boosting detection accuracy and inference speed. The updated version features an advanced decoder design that integrates cross-attention mechanisms with dynamic query generation, allowing the network to more effectively localize regions of interest. Additionally, RT-DETR-v2 adopts a multi-scale transformer architecture, which improves the model's ability to detect objects across a wide range of sizes by processing features from multiple resolution levels. Enhanced training protocols, including refined data augmentation techniques and modified loss functions, further contribute to mitigating class imbalance and localization challenges. These innovations are promising for delivering real-time performance without compromising precision in complex, real-world environments.

Hence, in this study, the seven types of real-time object detectors were evaluated, including YOLOv8, YOLOv9, YOLOv10, RT-DETR v11, YOLOv12, RT-DETR-v1, and RT-DETR v2 for blueberry detection. For each detector type, models at different scales (i.e., nano, small, medium, large, and extra-large for the YOLO series; and backbone variations such as ResNet-18, ResNet-34, and ResNet-50 for the RT-DETR series)   were trained and benchmarked for blueberry detection in terms of accuracy and inference time, using the dataset acquired in real orchards, resulting in a total suite of 36 model   variants included in the benchmark. The open-source software packages for implementing the selected detector are summarized in Table~\ref{tab:table2}.

\begin{table}[ht]
\caption{Overview of object detection models: URLs and corresponding references for the YOLO and RT-DETR model implementations used in this study.}
\label{tab:table2}
    \centering
    \resizebox{0.5\textwidth}{!}{%
        \begin{tabular}{lcc}
            \toprule
            Models & URL & References \\
            \midrule
            YOLOv8     & https://github.com/ultralytics/ultralytics    & \cite{jocher2023}    \\
            YOLOv9       & https://github.com/WongKinYiu/yolov9   & \cite{wang2024yolov9}      \\
            YOLOv10     & https://github.com/THU-MIG/yolov10    & \cite{wang2024yolov10}     \\
            YOLOv11     &https://github.com/ultralytics/ultralytics  & \cite{jocher2025yolo11edge}   \\
            YOLOv12    &https://github.com/sunsmarterjie/yolov12     & \cite{tian2025yolov12}  \\ 
            RT-DETR-v1    &https://github.com/lyuwenyu/RT-DETR & \cite{zhao2024detrs}   \\
            RT-DETR-v2    &https://github.com/lyuwenyu/RT-DETR  & \cite{lv2024rt}   \\
            \bottomrule
        \end{tabular}
    }
\end{table}

\subsection{Semi-Supervised Learning (SSL) for Enhanced Blueberry Detection}
SSL that aims to exploit unlabeled data for enhanced performance of supervised modeling has gained growing interest \citep{yang2022survey}, since manually labeling data, such as blueberry canopy images, can be a resource-intensive and time-consuming process. For object detection, SSL generally implements a teacher-student learning framework by first training a detection model (teacher) using labeled data, then generating bounding box predictions and pseudo labels for unlabeled data, and finally utilizing them alongside the labeled data to retrain the detector (student) \citep{deng2025semi}. This approach can address the challenge of both the limited labeled data and the abundant unlabeled images, with the potential to enhance detection performance. In this study, SSL was particularly justified for blueberry detection, as the newly collected dataset (Section~\ref{sec:2.1}). With such a large dataset, SSL offers a powerful approach to leverage the remaining unlabeled images by generating pseudo labels through the teacher model and incorporating them during the student model training. SSL can substantially reduce the annotation workload while improving detection robustness and scalability, ensuring the trained detector generalizes more effectively to real orchard environments.

\subsubsection{Unbiased Mean Teacher for YOLO Detectors}
For the YOLOv8, YOLOv9, YOLOv10, YOLOv11, and YOLOv12 detectors, semi-supervised learning was implemented through an Unbiased Mean Teacher (UMT) framework \citep{deng2021unbiased}, which combines the stability of the Mean Teacher \citep{tarvainen2017mean} paradigm with bias-mitigation strategies adapted from Unbiased Teacher \citep{liu2021unbiased}. By combining the simplicity and model-agnostic design of Mean Teacher with the bias-correction techniques of Unbiased Teacher, this hybrid Unbiased Mean Teacher framework is particularly well-suited for one-stage detectors such as the YOLO family. By leveraging both labeled and unlabeled data, the UMT framework enables the YOLO detectors to minimize reliance on exhaustive manual annotation while still improving overall performance. The teacher model, maintained as an exponential moving average (EMA) of the student, generates pseudo-labels for unlabeled images, while bias-reduction strategies, such as confidence-based filtering, mitigate the propagation of noisy or skewed pseudo-annotations. This hybrid approach not only stabilizes training but also enhances generalization, making it well-suited for real-time detection scenarios where data imbalance and limited annotations are common challenges.

\subsubsection{Semi-DETR for Detection Transformers}
For the RT-DETR v1 and RT-DETR v2 detectors, semi-supervised learning was implemented using the Semi-DETR framework \citep{zhang2023semi}, which was specifically developed to extend teacher–student learning to transformer-based detectors. Unlike generic teacher–student approaches, Semi-DETR is tightly coupled with the query-based design and Hungarian matching mechanism of DETR, allowing pseudo-labels from the teacher to be optimally assigned to queries in the student. This makes Semi-DETR particularly suitable for RT-DETR models, which rely on efficient query-based decoding for real-time detection. The framework introduces three key refinements: (1) confidence-aware pseudo-label filtering to reduce noise, (2) adaptive weighting of pseudo-labels based on prediction reliability, and (3) consistency regularization across weakly and strongly augmented inputs, ensuring prediction stability under perturbations. These characteristics allow Semi-DETR to maintain the efficiency of RT-DETR v1 and v2 while enhancing robustness and generalization by exploiting large volumes of unlabeled data.

\subsection{Experimentation}
\begin{figure*}[!ht]
 \centering
 \includegraphics[width=1.0\textwidth]{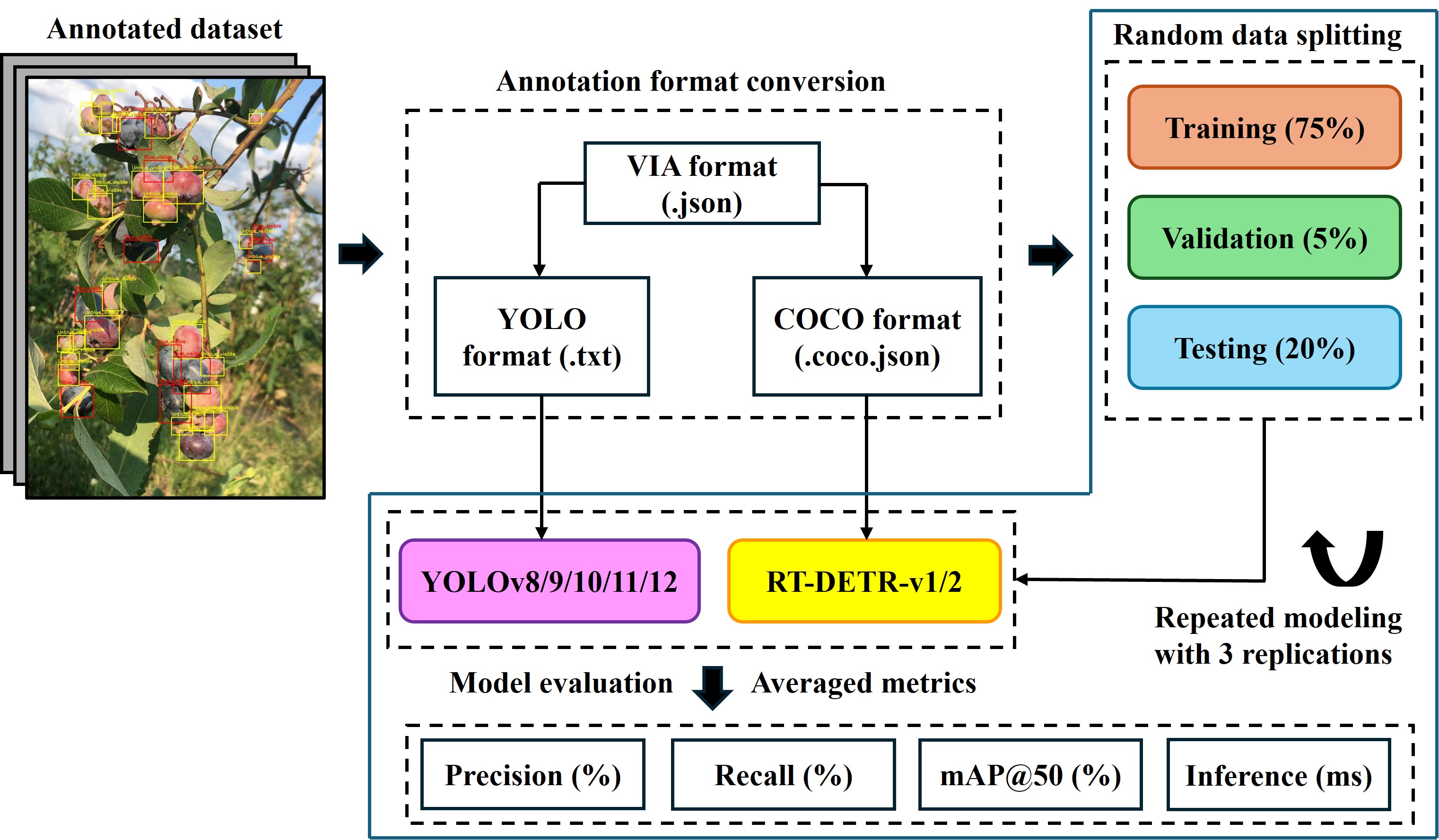}
 \caption{The pipeline flowchart of blueberry detection by YOLO and RT-DETR object detectors.}
 \label{fig:fig3}
\end{figure*}

Figure~\ref{fig:fig3} outlines the workflow pipeline for blueberry detection without SSL. Initially, the corresponding annotation files were converted into two distinct formats: YOLO format for YOLO-based detectors and COCO format for RT-DETR models. Following this, the dataset was randomly split into training (75\%), validation (5\%), and testing (20\%) subsets to ensure a balanced distribution for robust model development and unbiased evaluation. The models are then trained using their respective architectures and pre-trained weights and assessed against performance metrics (Session~\ref{sec:sec2.5}) to determine both detection accuracy and inference efficiency on the testing data.  All models were trained for 200 epochs, which were adequate for model convergence, using a uniform batch size of 32 across experiments. The raw images were resized to the resolution of 1024 × 1024 pixels for model input, which was empirically selected to preserve the fine-grained features of blueberries while ensuring real-time detection efficiency. Higher resolutions could be beneficial for small-object detection but substantially increase both training and inference times \citep{deng2024detection}.

For the YOLO-based detectors, additional hyperparameters included an initial learning rate of 0.01, momentum of 0.9, and a weight decay of 0.0005, along with a cosine annealing scheduler that included a 5-epoch warm-up phase to facilitate gradual learning rate reduction and fine-tuning in later epochs. Built-in data augmentation techniques (e.g., random horizontal flipping, rotations within a ±10° range, and color jittering) were automatically applied to enrich the training dataset and improve the detection of small objects, such as blueberries. The RT-DETR models were optimized using the Adam optimizer with an initial learning rate of 0.0001 and were trained on raw images without any additional data augmentation, relying on their transformer-based architecture to learn robust feature representations. To obtain a reliable estimate of model performance, replicated hold-out validations with three replications were conducted as done previously (Deng et al., 2024). The averaged metrics for precision, recall, and mean average precision (mAP@50) (Section~\ref{sec:sec2.5.1}) were calculated to assess blueberry detection accuracy for each model.

\begin{figure*}[!ht]
 \centering
 \includegraphics[width=1.0\textwidth]{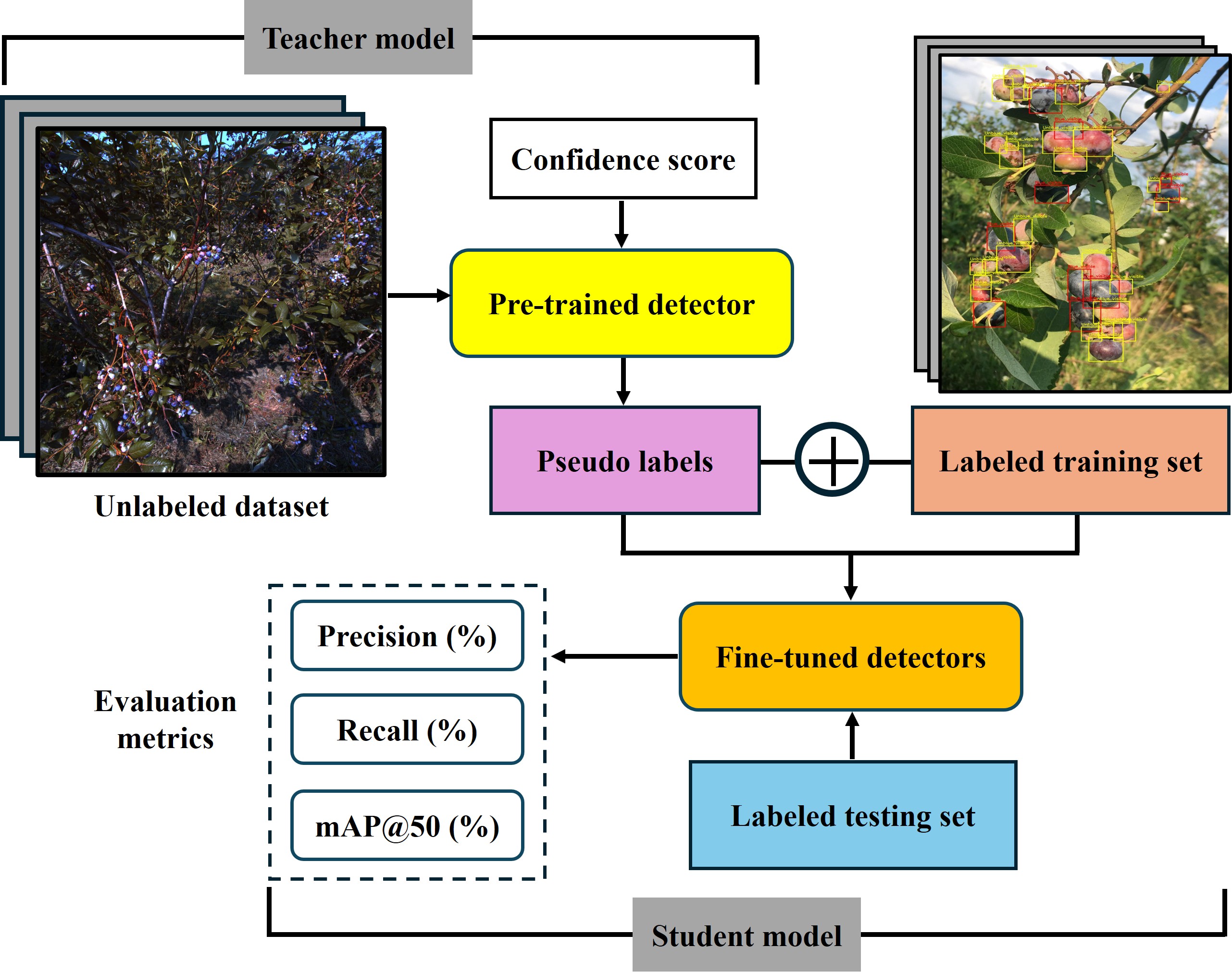}
 \caption{The flowchart of blueberry semi-supervised learning-based blueberry detection.}
 \label{fig:fig4}
\end{figure*}

For semi-supervised learning, YOLO detectors and RT-DETR v1/v2 were trained under a teacher–student framework, as shown in Figure~\ref{fig:fig4}. In both cases, the teacher model was updated as the EMA of the student with a momentum coefficient of $\alpha$ = 0.999, generating pseudo-labels on unlabeled images. Pseudo-labels were retained only if they exceeded a confidence threshold that was set at 0.9 initially and gradually relaxed to 0.7 after 20 epochs, and their confidence scores were used as weight when contributing to the loss. The unsupervised loss coefficient ($\lambda_u$) was ramped from 0.1 to 0.5 over the first 10 epochs to reduce the effect of noisy pseudo-labels in the early stages of training. Consistency regularization was applied by using weak augmentations for teacher predictions and strong augmentations for student training, with YOLO models using Mosaic, MixUp, random scaling, and color jittering, while RT-DETR models used random cropping, scaling, flipping, and color jittering. Framework-specific differences were preserved: YOLO detectors followed the standard YOLO loss formulation (classification, bounding box regression, and objectness), whereas RT-DETR used its transformer-based loss, including Hungarian matching for query assignment, classification and bounding box regression (L1 and GIoU), and auxiliary losses across decoder layers. In both model families, evaluation was performed on the same testing set used in the fully supervised experiments, ensuring direct comparability of results across supervised and semi-supervised settings.

The modeling experiments were performed on a high-performance workstation equipped with an Intel i9-10900X CPU (256 GB RAM) and a high-end graphics processing unit (GPU) (NVIDIA RTX A6000, 48 GB RAM), which provided the necessary computational power for accelerated model training. The software programs for modeling were developed in the environment of Python 3.8. Key libraries included OpenCV (version   4.8.0) for image processing, xml.etree.ElementTree (built-in with Python 3.8) for parsing annotation files, and the Ultralytics library (version 8.0.196) for implementing the YOLO detectors. Model training and inference were conducted using Pytorch (version 2.0.1) deep learning framework.   Additional routine libraries such as NumPy (1.24.4) for numerical computations, Pandas (1.5.3) for data handling, and Matplotlib (3.7.1) for visualization were also employed, ensuring a robust and reproducible experimental environment.

\subsection{Performance Evaluation Metrics}
\label{sec:sec2.5}
\subsubsection{Detection Accuracy}
\label{sec:sec2.5.1}
The blueberry detection accuracy was assessed using common metrics, including precision, recall, and mAP@50, as done previously \citep{deng2024detection}. These metrics are calculated based on three independent experimental replicates with the following equations:

\begin{equation} \label{eqn:p}
Precision(\%) = \frac{\#TP}{\#TP + \#FP}\times100\%,
\end{equation}

\begin{equation} \label{eqn:r}
Recall(\%) = \frac{\#TP}{\#TP + \#FN}\times100\%,
\end{equation}

\begin{equation} \label{eqn:map50}
mAP@50 (\%) = \frac{1}{N} \sum_{i=1}^{N} (AP)_i \times 100\%
\end{equation}

Where \#TP, \#FP, \#TN, and \#FN represent the numbers of true positive, false positive, true negative, and false negative detection for each class (“Blue” and “Unblue”), respectively, and N=2, representing the number of blueberry maturity classes considered in this study.

It is noted that given the two classes of blueberries, “Blue” (ripe blueberries) and “Unblue” (unripe blueberries), a true positive (TP) was defined as a predicted bounding box that not only overlapped with a ground truth annotation of the same class above the IoU threshold (0.5) but also matched the correct label. For example, a prediction labeled as “Blue” that corresponded to a ground truth blue fruit was counted as a true positive, while misclassifying a blue fruit as “Unblue” (or vice versa) was treated as a false positive (FP) error. Precision, defined as the ratio of TP blueberry detections to all detected instances, indicates the extent to which FPs are minimized, ensuring that most detections correspond to actual blueberries rather than background noise. Recall, calculated as the ratio of TPs to the sum of TPs and false negatives within each class, measures the ability to capture all annotated instances of that class rather than all blueberries in general. Both precision and recall were first obtained class-wise and then averaged across the two categories. Compared to Precision and Recall, the mAP@50, which is widely considered a primary metric for object detection, was computed across varying confidence thresholds and averaged over both maturity classes, providing a comprehensive evaluation of overall detection accuracy.

\subsubsection{Model Complexity and Inference Time}
\label{sec:sec2.5.2}
The model complexity was evaluated in terms of giga floating-point operations (GFLOPs), which represents the number of required arithmetic operations for a single forward pass on an image of fixed resolution. Lower GFLOPs generally correspond to reduced computational demand and better real-time performance on resource-constrained hardware. Together, these metrics provide a comprehensive view of the detection accuracy and computational cost, enabling informed decisions for selecting models that achieve optimal performance in targeted application scenarios.

To evaluate computational efficiency and assess the practicality of deploying the models in different operational environments, three more metrics were considered: inference time, training time, and model complexity. Training time was recorded for the full training cycle of each model, from initialization to convergence, to quantify the computational resources required for model development. Longer training times generally indicate higher computational demands, which may influence the feasibility of frequent fine-tuning or large-scale model experimentation. Inference time was measured as the average processing time per image over the test set. Two hardware configurations were used: (1) a desktop computer equipped with an Intel i9-11900 CPU (64 GB RAM) and an NVIDIA GeForce RTX 4060 Ti GPU (16 GB RAM), and (2) a Jetson Orin 64 GB computer (NVIDIA, Santa Clara, CA) that represents real-time field deployment conditions.

\section{Results and Discussion}
\subsection{Fully Supervised Learning}
\subsubsection{Detection Accuracy}
Table~\ref{tab:table3} summarizes the detection accuracy (precision, recall, and mAP@50) of the YOLO models (v8–v12) and RT-DETR detectors (v1 and v2). The results highlight performance differences both within each family and across detector architectures.

Within the YOLO series, YOLOv8 established a strong foundation, with the lightweight YOLOv8n achieving 88.9\% precision, 88.7\% recall, and 90.9\% mAP@50. Larger variants such as YOLOv8m and YOLOv8l exceeded 90\% in precision and recall, with mAP@50 values of 91.6\% and 92.3\%, respectively. The YOLOv9 series followed a similar pattern, with YOLOv9s producing 89.7\% mAP@50 and the largest variant, YOLOv9e, reaching 92.7\% while maintaining precision and recall above 90\%. By contrast, YOLOv10 underperformed relatively to its predecessors, with YOLOv10n achieving 88.5\% mAP@50 and the largest model, YOLOv10x, only reaching 89.8\%. The YOLOv11 series performed better than YOLOv10, where precision and recall approached 91\% across its larger models (mid to extra-large), and particularly YOLOv11x achieved mAP@50 of 92.3\%. The YOLOv12 series offered further improvements over YOLOv11x at all model scales, with precision and recall consistently above 91\% in mid- to extra-large variants, delivering the highest mAP@50 of 93.3\% by YOLOv12m among all the YOLO models, thereby establishing YOLOv12 as the most accurate version in the family.

For the RT-DETR detectors, in the v1 series, RT-DETR-R18 attained 87.3\% mAP@50 with precision and recall of around 85\%, and the larger model appeared to produce better accuracy. For instance, RT-DETR- HGNetv2-X resulted in the best accuracy among the v1 models, with 91.3\% mAP@50 and both precision and recall approaching 90\%. The RT-DETR-v2 series outperformed the v1 series, with mAP@50 consistently exceeding 90\%. The RT-DETR-v2-S achieved 90.6\% mAP@50, and remarkably, RT-DETR-v2-X reached 93.6\% mAP@50, the highest accuracy observed across all examined detectors, accompanied by precision and recall above 92\%.

\begin{figure*}[!ht]
 \centering
 \includegraphics[width=1.0\textwidth]{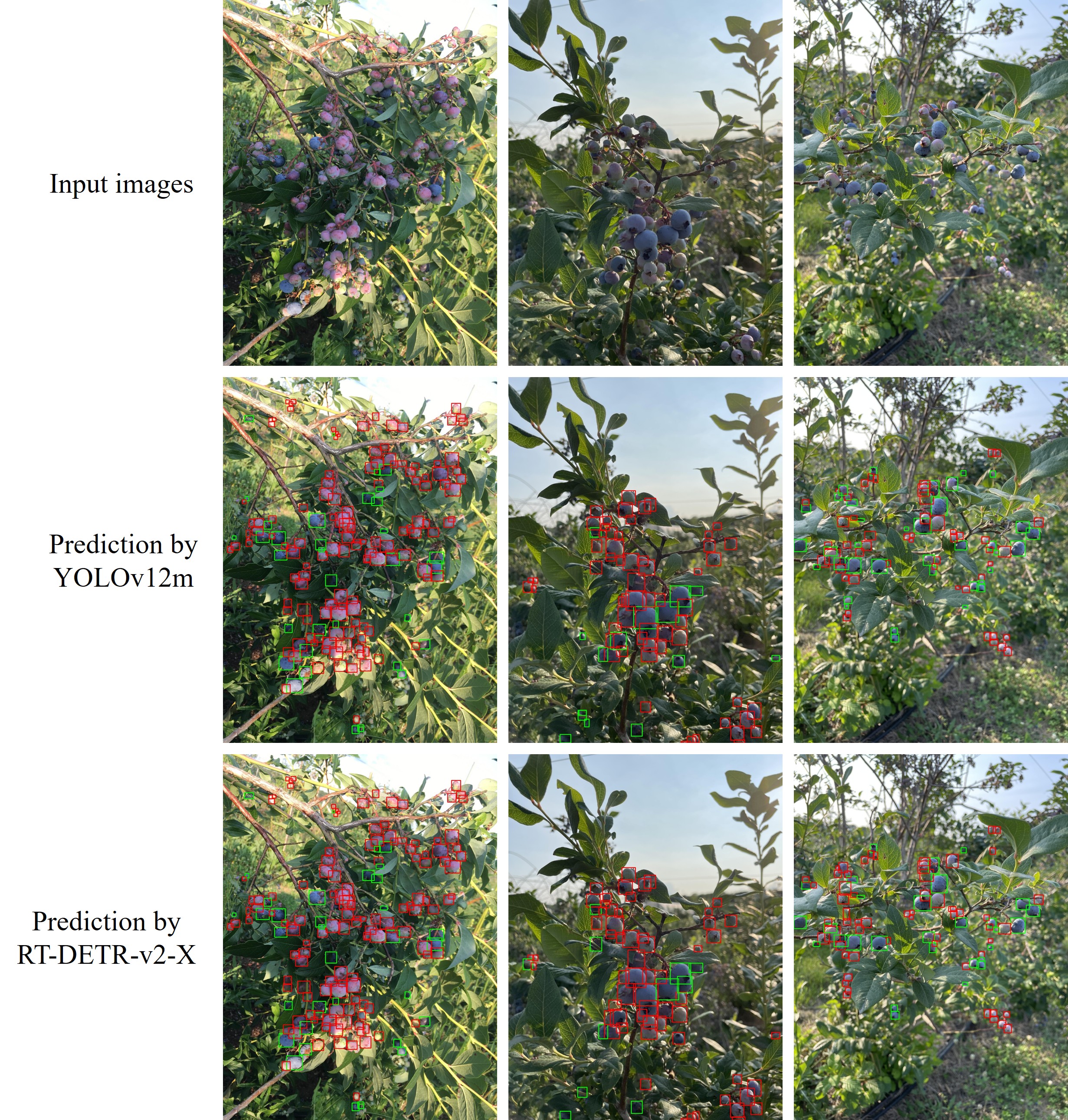}
 \caption{Example blueberry detection by YOLOv12m and RT-DETR-v2-X.}
 \label{fig:fig5}
\end{figure*}

Overall, all YOLO models showed competitive precision-recall balances, and particularly YOLOv8, YOLOv11, and YOLOv12 offered competitive performance, making them well-suited for applications demanding reliable detection with scalable complexity. The performance of YOLOv12m (93.3\% mAP@50) and RT-DETR-v2-X (93.6\% mAP@50) was on par with each other, with only a 0.3\% difference in mAP@50. Figure~\ref{fig:fig5} shows example prediction results, where both models successfully detect the vast majority of blueberries across varying illumination and occlusion conditions. Despite the importance of detection accuracy, the choice of detectors for practical application should not be solely based on this metric; it is also important to factor the model complexity and inference time into decision-making, as presented below for the examined detectors.

\begin{sidewaystable*}[htb!]
\centering
\caption{Performance comparison of object detection models. Summary of the computational complexity (GFLOPs), detection accuracy (mAP@50), and inference times for different real-time detection models [i.e., YOLO (v8-v12) and RT-DETR (v1-v2) families].}
\label{tab:table3}
\renewcommand{\arraystretch}{1.5}
\begin{tabular}{llcccccc}
\toprule
\textbf{Models} & \textbf{Submodels} & \textbf{GFLOPs} & \textbf{Precision (\%)} & \textbf{Recall (\%)} & \textbf{mAP@50 (\%)} & \multicolumn{2}{c}{\textbf{Inference Time (ms)}} \\ 
\cmidrule(lr){7-8}
& & & & & & \textbf{Work station} & \textbf{Jetson Orin} \\
\midrule
\multirow{5}{*}{YOLOv8}  & YOLOv8n  & 8.2   & 88.9$\pm$0.5 & 88.7$\pm$0.3 & 90.9$\pm$0.4 &  51.1 & 122.6 \\
                         & YOLOv8s  & 28.6  & 89.2$\pm$0.5 & 89.1$\pm$0.4 & 90.8$\pm$0.9 & 102.7 & 246.5 \\
                         & YOLOv8m  & 79.1  & 90.4$\pm$0.5 & 90.0$\pm$0.3 & 91.6$\pm$0.2 & 366.2 & 878.9 \\
                         & YOLOv8l  & 165.4 & 90.4$\pm$0.5 & 90.2$\pm$0.6 & 92.3$\pm$0.5 & 387.6 & 939.2 \\
                         & YOLOv8x  & 257.4 & 89.8$\pm$0.3 & 90.0$\pm$0.5 & 91.5$\pm$0.8 & 587.0 & 1467.5 \\
\midrule
\multirow{4}{*}{YOLOv9}  & YOLOv9s  & 39.6  & 87.7$\pm$0.4 & 87.4$\pm$0.5 & 89.7$\pm$1.6 & 183.2 & 444.6 \\
                         & YOLOv9m  & 132.4 & 91.0$\pm$0.3 & 90.5$\pm$0.6 & 92.2$\pm$0.5 & 397.9 & 959.1 \\
                         & YOLOv9c  & 238.9 & 90.2$\pm$0.6 & 89.6$\pm$0.5 & 91.4$\pm$0.5 & 553.5 & 1345.3 \\
                         & YOLOv9e  & 244.9 & 91.2$\pm$0.5 & 90.8$\pm$0.4 & 92.7$\pm$0.8 & 616.4 & 1533.4 \\
\midrule
\multirow{6}{*}{YOLOv10} & YOLOv10n & 15.5  & 87.2$\pm$0.4 & 86.4$\pm$0.6 & 88.5$\pm$2.3 & 121.2 & 297.1 \\
                         & YOLOv10s & 44.8  & 89.3$\pm$0.3 & 88.6$\pm$0.5 & 90.4$\pm$1.1 & 208.3 & 499.2 \\
                         & YOLOv10m & 154.9 & 88.8$\pm$0.4 & 88.2$\pm$0.5 & 90.2$\pm$0.4 & 441.7 & 1104.5 \\
                         & YOLOv10b & 265.7 & 87.8$\pm$0.4 & 87.9$\pm$0.6 & 89.6$\pm$0.3 & 586.1 & 1430.0 \\
                         & YOLOv10l & 304.1 & 87.8$\pm$0.5 & 87.2$\pm$0.4 & 89.3$\pm$0.9 & 652.1 & 1477.6 \\
                         & YOLOv10x & 355.8 & 88.8$\pm$0.5 & 87.7$\pm$0.3 & 89.8$\pm$1.7 & 726.9 & 1587.8 \\
\midrule
\multirow{5}{*}{YOLOv11} & YOLOv11n & 6.3   & 87.3$\pm$0.6 & 86.9$\pm$0.4 & 88.5$\pm$0.3 &  89.6 & 134.3 \\
                         & YOLOv11s & 21.3  & 89.2$\pm$0.4 & 88.6$\pm$0.4 & 91.1$\pm$0.6 & 143.6 & 214.2 \\
                         & YOLOv11m & 67.7  & 90.1$\pm$0.3 & 90.1$\pm$0.5 & 91.9$\pm$0.4 & 225.1 & 458.6 \\
                         & YOLOv11l & 86.6  & 90.8$\pm$0.5 & 89.7$\pm$0.6 & 91.9$\pm$0.8 & 488.9 & 872.9 \\
                         & YOLOv11x & 194.4 & 90.9$\pm$0.5 & 90.7$\pm$0.3 & 92.3$\pm$0.2 & 694.1 & 1387.4 \\
\midrule
\multirow{5}{*}{YOLOv12} & YOLOv12n & 6.3   & 88.1$\pm$0.5 & 88.1$\pm$0.5 & 89.6$\pm$0.7 & 105.9 & 254.3 \\
                         & YOLOv12s & 21.2  & 89.9$\pm$0.5 & 90.0$\pm$0.4 & 91.8$\pm$2.1 & 259.6 & 621.1 \\
                         & YOLOv12m & 67.1  & 91.6$\pm$0.4 & 91.5$\pm$0.4 & 93.3$\pm$1.5 & 478.1 & 815.6 \\
                         & YOLOv12l & 88.6  & 90.7$\pm$0.5 & 90.2$\pm$0.3 & 92.2$\pm$0.9 & 567.8 & 1155.8 \\
                         & YOLOv12x & 198.5 & 91.6$\pm$0.4 & 90.3$\pm$0.4 & 92.8$\pm$1.3 & 654.4 & 1395.5 \\
\bottomrule
\end{tabular}%
\end{sidewaystable*}

\begin{sidewaystable*}[htb!]
\ContinuedFloat
\centering
\captionsetup{labelformat=empty}
\caption{Table 3 (continued). Performance comparison of object detection models. Summary of the computational complexity (GFLOPs), detection accuracy (mAP@50), and inference times for different real-time detection models [i.e., YOLO (v8-v12) and RT-DETR (v1-v2) families]}
\renewcommand{\arraystretch}{1.5}
\begin{tabular}{llcccccc}
\toprule
\textbf{Models} & \textbf{Submodels} & \textbf{GFLOPs} & \textbf{Precision (\%)} & \textbf{Recall (\%)} & \textbf{mAP@50 (\%)} & \multicolumn{2}{c}{\textbf{Inference Time (ms)}} \\ 
\cmidrule(lr){7-8}
& & & & & & \textbf{Work station} & \textbf{Jetson Orin} \\
\midrule
\multirow{7}{*}{RT-DETR-v1} 
  & RT-DETR-R18        & 60.5  & 85.5$\pm$0.3 & 84.9$\pm$0.6 & 87.3$\pm$0.6 &  55.2 & 131.4 \\
  & RT-DETR-R34        & 92.3  & 87.3$\pm$0.3 & 87.1$\pm$0.4 & 89.2$\pm$0.9 &  89.7 & 217.4 \\
  & RT-DETR-R50-m      & 100.8 & 88.0$\pm$0.3 & 88.3$\pm$0.5 & 89.9$\pm$1.1 & 186.8 & 466.2 \\
  & RT-DETR-R50        & 136.1 & 89.0$\pm$0.5 & 88.7$\pm$0.5 & 90.2$\pm$1.5 & 355.1 & 878.4 \\
  & RT-DETR-R101       & 259.6 & 89.8$\pm$0.4 & 89.2$\pm$0.4 & 91.1$\pm$2.3 & 564.9 & 1007.1 \\
  & RT-DETR-HGNetv2-L  & 110.8 & 89.5$\pm$0.5 & 88.5$\pm$0.6 & 90.8$\pm$1.8 & 489.5 & 936.7 \\
  & RT-DETR-HGNetv2-X  & 234.5 & 89.9$\pm$0.6 & 89.5$\pm$0.6 & 91.3$\pm$0.5 & 752.3 & 1733.9 \\
\midrule
\multirow{4}{*}{RT-DETR-v2} 
  & RT-DETR-v2-S       & 60.6  & 89.1$\pm$0.4 & 89.0$\pm$0.4 & 90.6$\pm$1.7 & 264.3 & 455.3 \\
  & RT-DETR-v2-M       & 100.4 & 89.7$\pm$0.4 & 89.9$\pm$0.4 & 91.5$\pm$1.1 & 385.4 & 568.1 \\
  & RT-DETR-v2-L       & 136.9 & 90.4$\pm$0.5 & 90.1$\pm$0.4 & 92.4$\pm$0.8 & 597.6 & 763.9 \\
  & RT-DETR-v2-X       & 259.1 & 92.4$\pm$0.4 & 92.1$\pm$0.4 & 93.6$\pm$2.4 & 834.1 & 1973.4 \\
\bottomrule
\end{tabular}%
\end{sidewaystable*}

\subsubsection{Detection Speed}
\begin{figure*}[!ht]
 \centering
 \includegraphics[width=1.0\textwidth]{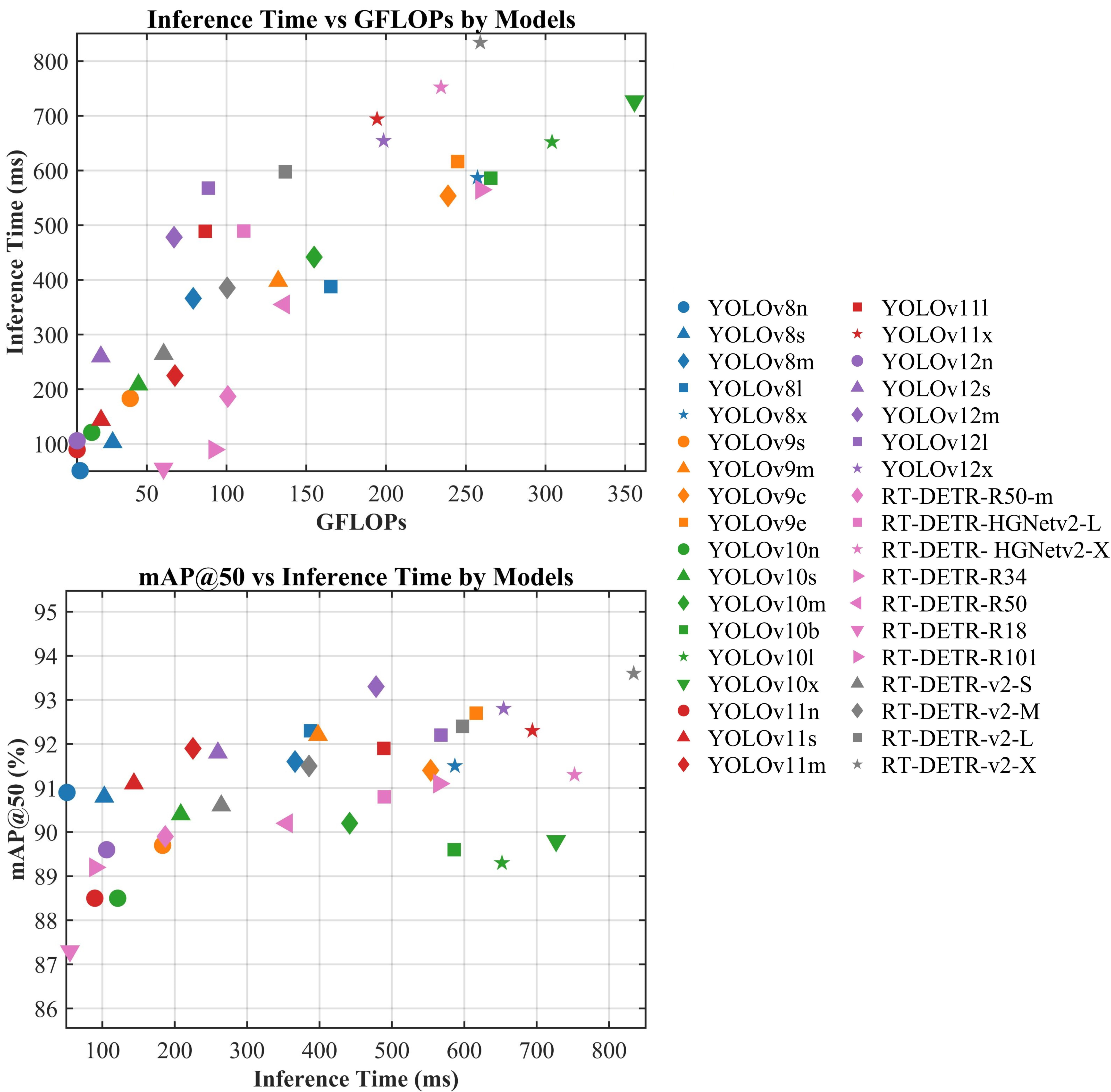}
 \caption{Scatter plots of inference time versus GFLOPs (giga floating-point operations) and mAP@50 versus inference time in blueberry detection by all the YOLO and RT-DETR model variants.}
 \label{fig:fig6}
\end{figure*}

The detection speed is associated with the model complexity and computation costs as measured by the GLOPs. Figure~\ref{fig:fig6} shows that the inference times increase positively with the GLOPs for a given model series, although their relationship appears to be nonlinear. A trade-off between detection accuracy and speed exists across different YOLO and RT-DETR model variants (Figure~\ref{fig:fig6}). Smaller models such as YOLOv8n, YOLOv8s, and RT-DETR-R18 achieve faster inference times (<150 ms) but at the cost of slightly lower accuracy (around 89–91\%). Larger and more complex architectures, including YOLOv12m, YOLOv12l, RT-DETR-v2-L, and RT-DETR-v2-X, deliver higher accuracy (above 92–94\%) but require substantially longer inference times (500–800 ms). The mid-sized models (e.g., YOLOv9m, YOLOv11m, and RT-DETR-R50) appear to offer a good balance, maintaining competitive accuracy while keeping inference within 300–500 ms. These results highlight the general accuracy–efficiency trade-off, as noted in other studies \citep{dang2023yoloweeds,le2024benchmarking}, for practical model deployment when computing resources are a constraint.

\subsection{Semi-Supervised Blueberry Detection}
Table~\ref{tab:table4} presents the comparison between the original, fully-supervised detectors and their SSL fine-tuned counterparts for blueberry detection. Overall, despite mixed effects of SSL, the majority of the YOLO (14 out of 25) and RT-DETR (8 out of 11) models attained performance gains over their supervised baselines, which varied with model. In some cases, SSL enhanced accuracy substantially, while in others it produced only marginal changes or even slight declines.

For YOLOv8, the smaller variants, YOLOv8n and YOLOv8s, saw decreases of 1.3 and 0.9 percentage points in mAP@50, respectively, probably because the two lightweight models may be less capable of leveraging the additional pseudo-labeled data and more susceptible to noise. In contrast, the mid-sized YOLOv8m and YOLOv8l remained more stable, and the largest variant, YOLOv8x, benefited considerably from SSL, improving by 1.7 points. Likewise, the two larger versions of the YOLOv9 series, YOLOv9c and YOLOv9e, gained 0.9 and 1.5 points, respectively, with the best accuracy of 94.2\% by the latter among all the YOLO models. The YOLOv10 series showed the most consistent benefits at all scales, with YOLOv10n improving by 1.7 points and YOLOv10x gaining the largest boost of 2.9 points among all the YOLO models. By contrast, the YOLOv11 series showed only marginal benefits, with small gains in some variants but modest declines in others. The YOLOv12 series was similarly inconsistent, with improvements in certain models but decreases or negligible changes in others. These outcomes highlight that newer architecture does not necessarily guarantee greater compatibility with SSL, possibly because the UMT framework was not specifically optimized for these detectors. As a result, the interaction between pseudo-label generation and architectural design may have limited the benefits for certain models.

The RT-DETR detectors revealed a similar mixed pattern due to SSL. In the v1 series, for instance, RT-DETR-R18 and R50 improved substantially by 1.6 and 1.5 points, respectively, while RT-DETR-R34 and R101 experienced slight declines. By contrast, the RT-DETR-v2 series displayed a more consistent benefit, with the RT-DETR-v2-S, v2-L, and v2-X improving by 1.1–1.5 points, and notably, the latter (RT-DETR-v2-X) achieved the best accuracy of 94.8\% among all the examined models.

Overall, these results indicate that the effectiveness of SSL depends not only on model capacity but also on how well the framework aligns with the detector architecture. In the YOLO family, the UMT framework delivered clear benefits in some cases, most notably for YOLOv10, but the outcomes were inconsistent across versions. This suggests that while UMT can enhance convolution-based detectors, its pseudo-labeling and consistency regularization strategies may not fully exploit the architectural refinements introduced in later YOLO models such as v11 and v12. The variability observed in YOLO highlights that newer architectures do not automatically guarantee greater compatibility with SSL, underscoring the importance of framework–architecture co-design.

In contrast, the RT-DETR family showed overall more stable gains under the Semi-DETR framework. The v1 series achieved improvements in most models, with notable gains in R18 and R50, while only R34 and R101 registered slight decreases. The v2 series further strengthened this trend, with three out of four models improving by 1.1–1.5 mAP@50 and only v2-M showing a marginal decline. These consistent improvements in the RT-DETR family can be attributed to the fact that Semi-DETR was specifically designed for DETR-based architecture, allowing it to integrate more effectively with transformer-style detection pipelines. This alignment likely explains why RT-DETR models, particularly in the v2 series, benefited more reliably from SSL compared with YOLO.

\begin{table*}[htbp]
\centering
\caption{Impact of semi-supervised learning (SSL) on blueberry detection accuracy comparing YOLO and RT-DETR models with and without SSL fine-tuning, with the performance differences indicated in parentheses.}
\label{tab:table4}
\renewcommand{\arraystretch}{1.3}
\begin{tabular}{llccc}
\toprule
\textbf{Models} & \textbf{Submodels} & \textbf{Precision (\%)} & \textbf{Recall (\%)} & \textbf{mAP@50\_SSL (\%)} \\
\midrule
\multirow{5}{*}{YOLOv8} 
 & YOLOv8n  & 88.8 & 88.3 & 89.6 \textcolor{green!60!black}{(-1.3)} \\
 & YOLOv8s  & 88.7 & 89.4 & 89.9 \textcolor{green!60!black}{(-0.9)} \\
 & YOLOv8m  & 89.1 & 90.7 & 91.5 \textcolor{green!60!black}{(-0.1)} \\
 & YOLOv8l  & 89.6 & 90.4 & 92.1 \textcolor{green!60!black}{(-0.2)} \\
 & YOLOv8x  & 91.9 & 92.5 & 93.2 \textcolor{red}{(+1.7)} \\
\midrule
\multirow{4}{*}{YOLOv9} 
 & YOLOv9s  & 88.6 & 89.0 & 90.1 \textcolor{red}{(+0.4)} \\
 & YOLOv9m  & 89.1 & 89.9 & 90.8 \textcolor{green!60!black}{(-1.4)} \\
 & YOLOv9c  & 89.9 & 91.8 & 92.3 \textcolor{red}{(+0.9)} \\
 & YOLOv9e  & 93.8 & 94.5 & 94.2 \textcolor{red}{(+1.5)} \\
\midrule
\multirow{6}{*}{YOLOv10} 
 & YOLOv10n & 88.1 & 89.8 & 90.2 \textcolor{red}{(+1.7)} \\
 & YOLOv10s & 89.5 & 90.1 & 91.1 \textcolor{red}{(+0.7)} \\
 & YOLOv10m & 89.7 & 89.9 & 90.8 \textcolor{red}{(+0.6)} \\
 & YOLOv10b & 89.1 & 89.2 & 90.4 \textcolor{red}{(+0.8)} \\
 & YOLOv10l & 89.1 & 89.6 & 90.6 \textcolor{red}{(+1.3)} \\
 & YOLOv10x & 91.8 & 92.5 & 92.7 \textcolor{red}{(+2.9)} \\
\midrule
\multirow{5}{*}{YOLOv11} 
 & YOLOv11n & 88.7 & 89.3 & 88.9 \textcolor{red}{(+0.4)} \\
 & YOLOv11s & 89.8 & 88.5 & 90.6 \textcolor{green!60!black}{(-0.5)} \\
 & YOLOv11m & 89.9 & 91.3 & 92.2 \textcolor{red}{(+0.3)} \\
 & YOLOv11l & 89.3 & 89.1 & 91.5 \textcolor{green!60!black}{(-0.4)} \\
 & YOLOv11x & 90.9 & 91.5 & 92.1 \textcolor{green!60!black}{(-0.2)} \\
\midrule
\multirow{5}{*}{YOLOv12} 
 & YOLOv12n & 87.6 & 87.6 & 88.9 \textcolor{green!60!black}{(-0.7)} \\
 & YOLOv12s & 90.7 & 91.4 & 92.1 \textcolor{red}{(+0.3)} \\
 & YOLOv12m & 90.4 & 92.7 & 93.2 \textcolor{green!60!black}{(-0.1)} \\
 & YOLOv12l & 91.7 & 92.3 & 92.6 \textcolor{red}{(+0.4)} \\
 & YOLOv12x & 90.8 & 91.5 & 92.4 \textcolor{green!60!black}{(-0.4)} \\
\midrule
\multirow{7}{*}{RT-DETR-v1} 
 & RT-DETR-R18       & 87.8 & 88.4 & 88.9 \textcolor{red}{(+1.6)} \\
 & RT-DETR-R34       & 85.8 & 87.5 & 88.6 \textcolor{green!60!black}{(-0.6)} \\
 & RT-DETR-R50-m     & 88.6 & 89.1 & 90.4 \textcolor{red}{(+0.5)} \\
 & RT-DETR-R50       & 89.8 & 90.0 & 91.7 \textcolor{red}{(+1.5)} \\
 & RT-DETR-R101      & 88.4 & 90.3 & 90.8 \textcolor{green!60!black}{(-0.3)} \\
 & RT-DETR-HGNetv2-L & 89.9 & 90.2 & 91.4 \textcolor{red}{(+0.6)} \\
 & RT-DETR-HGNetv2-X & 90.9 & 91.4 & 91.9 \textcolor{red}{(+0.6)} \\
\midrule
\multirow{4}{*}{RT-DETR-v2} 
 & RT-DETR-v2-S & 91.2 & 91.6 & 92.1 \textcolor{red}{(+1.5)} \\
 & RT-DETR-v2-M & 90.3 & 90.7 & 91.2 \textcolor{green!60!black}{(-0.3)} \\
 & RT-DETR-v2-L & 92.5 & 92.9 & 93.5 \textcolor{red}{(+1.1)} \\
 & RT-DETR-v2-X & 94.6 & 95.2 & 94.8 \textcolor{red}{(+1.2)} \\
\bottomrule
\end{tabular}
\end{table*}

\subsection{Discussion}

This study represents an important step forward in applying advanced deep learning techniques to blueberry detection. By systematically benchmarking state-of-the-art real-time detectors from the YOLO and RT-DETR families, the research provides a comprehensive evaluation of the performance of these model architectures, enabling informed model selection for orchard applications. Integrating SSL into the training pipeline is overall promising for enhanced model detection capabilities by leveraging unlabeled data, improving model robustness under diverse environmental conditions. The public release of a large, annotated blueberry dataset in this study establishes a new benchmark for agricultural computer vision research. This openly accessible resource will accelerate progress in blueberry detection, foster collaboration among researchers and practioners, and enable exploration of novel AI model architectures and training techniques to advance precision horticulture.

The findings of this work have the potential to transform blueberry production practices. Accurate and reliable detection of blueberries by the advanced AI-based detectors underpins essential downstream tasks such as yield estimation, maturity assessment, and selective harvesting. By implementing high-resolution images and advanced detection algorithms, the study addresses major challenges in distinguishing individual blueberries of varied maturity levels within dense, often occluded canopy structures. Automated counting and classification also minimize human error and reduce reliance on laborious manual work. The improved detection performance directly supports more precise yield predictions and optimized harvest scheduling, ultimately improving orchard management while reducing labor costs.

Our results demonstrate he value of SSL for agricultural applications where fully annotated datasets are costly to obtain. Fine-tuning with unlabeled data improved the generalization of high-capacity models and enhanced their robustness to seasonal and environmental variability, allowing growers to rely on consistent performance across diverse orchard conditions. Moreover, successfully adapting real-time object detection models to the unique challenges of fruit detection opens opportunities to extend the approach to visual recognition tasks for other crops. However, it is important to note that, despite overall positive impacts,  more in-depth, dedicated research into SSL techniques for enhanced blueberry detection is still needed to better leverage unlabeled data collected from diverse conditions. The unlabeled dataset acquired by the machine vision platform (Figure~\ref{fig:fig2}) possesses different characteristics compared to the labeled dataset collected with smartphones in different seasons, implying a distribution shift between the labeled and unlabeled datasets. This could have posed challenges to the UMT-based framework, which probably explains the reduced accuracy by some detectors in this study. A more extensive examination of advanced SSL techniques \citep{shehzadi2024, yang2022survey} will be beneficial for enhanced blueberry detection. 

Although the dataset in this study captures the diversity in blueberry bushes and orchard conditions, the acquired images do not offer a holistic view of the entire bushes, and the detection results are hence limited to localized branches. To enable better precision in orchard management, there is a need to image the entire blueberry bushes from multiple viewpoints, which requires a well-designed imaging platform equipped with multiple cameras beyond simply using handheld cameras for data collection. Research is underway \citep{mu2025platform} to address the need by developing an over-the-row machine vision platform that continuously scans the bushes for full-canopy fruit detection. 

\section{Conclusion}
\label{sec:conclu}
This study demonstrates that computer vision powered by AI can play a critical role in modern blueberry production. By developing a large, diverse dataset specifically curated for blueberry detection, comprising 661 canopy images and 85,879 manually annotated instances of ripe and unripe blueberries, the study has provided a solid database for evaluating state-of-the-art object detection models under realistic orchard conditions. Through benchmarking 36 advanced models from the YOLO (v8-v12) and RT-DETR (v1-v2) families, this work offers valuable insights into the trade-offs between accuracy, inference speed, and computational efficiency, enabling informed model selection. Particularly, YOLOv12m achieved the best accuracy with a mAP@50 of 93.3\% among the YOLO models, while RT-DETRv2-X obtained 93.6\% mAP@50, the highest in the RT-DETR family. Overall, the inference time varied with the model scale and complexity, and the mid-sized models appeared to offer a good balance between accuracy and speed. A UMT-based SSL framework was applied to fine-tune all the models with an additional set of 1,035 images acquired by a machine vision platform. Although results varied with accuracy changes ranging from slight decreases to gains of up to 2.9 percentage points, SSL demonstrated clear potential to enhance detection performance, with the RT-DETR-v2-X reaching the highest mAP@50 of 94.8\% after fine-tuning. 

Overall, this study advances blueberry detection by delivering both a benchmark dataset and a systematic evaluation of cutting-edge real-time detectors. The findings support the development of reliable, orchard-ready vision systems capable of blueberry detection and harvest maturity assessment, enabling downstream tasks such as yield estimation, harvesting scheduling, and precision harvesting. Future work will focus on refining SSL techniques and optimizing models for edge deployment while developing a full-canopy machine vision system for comprehensive fruit detection.

\section*{Acknowledgment}
This work was supported by the Hatch Multistate Programmatic Funding of Michigan State University AgBioResearch. The authors thank Mingjun Li for assisting in dataset curation and model training. 

\typeout{}
\bibliography{Benchmark_preprint}
\end{document}